\documentclass[11pt,letterpaper]{article}
\usepackage{aaai17}
\usepackage{amsmath}
\usepackage{times}
\usepackage{latexsym}
\usepackage{color}
\usepackage{booktabs}
\usepackage{subfig}
\usepackage{tikz}
\usepackage{fancybox}
\usetikzlibrary{bayesnet}
\usetikzlibrary{calc}

\newcommand{\secref}[1]{Section~\ref{sec:#1}}
\newcommand{\figref}[1]{Figure~\ref{fig:#1}}

\newcommand{\tabref}[1]{Table~\ref{tab:#1}}

\newcommand{\blank}{\underline{\hspace{.5cm}}}
\newcommand{\lexicalpredicate}[1]{\ensuremath{\textit{#1}}}
\newcommand{\formalpredicate}[1]{{\small \ensuremath{\textsc{#1}}}}
\newcommand{\entity}[1]{\ensuremath{\textsc{#1}}}
\newcommand{\prob}{\ensuremath{p}}
\newcommand{\pathstart}{\ensuremath{\langle}}
\newcommand{\pathend}{\ensuremath{\rangle}}

\newcommand\tikzmark[1]{%
  \tikz[remember picture,overlay]\node (#1) {};%
}

\title{Open-Vocabulary Semantic Parsing\\with both Distributional
Statistics and Formal Knowledge}

\author{Matt Gardner and Jayant Krishnamurthy\\
Allen Institute for Artificial Intelligence\\
Seattle, Washington, USA\\
{\tt \{mattg,jayantk\}@allenai.org}}

\date{}

\begin{document}

\maketitle

\begin{abstract}

  Traditional semantic parsers map language onto compositional, executable queries in a fixed
  schema.  This mapping allows them to effectively leverage the information contained in large,
  formal knowledge bases (KBs, e.g., Freebase) to answer questions, but it is also fundamentally
  limiting---these semantic parsers can only assign meaning to language that falls within the KB's
  manually-produced schema.  Recently proposed methods for open vocabulary semantic parsing
  overcome this limitation by learning execution models for arbitrary language, essentially using a
  text corpus as a kind of knowledge base.  However, all prior approaches to open vocabulary
  semantic parsing \emph{replace} a formal KB with textual information, making no use of the KB in
  their models.  We show how to combine the disparate representations used by these two approaches,
  presenting for the first time a semantic parser that (1) produces compositional, executable
  representations of language, (2) can successfully leverage the information contained in both a
  formal KB and a large corpus, and (3) is not limited to the schema of the underlying KB.  We
  demonstrate significantly improved performance over state-of-the-art baselines on an open-domain
  natural language question answering task.

\end{abstract}

\section{Introduction}

Semantic parsing is the task of mapping a phrase in natural language onto a formal query in some
fixed schema, which can then be executed against a knowledge
base (KB)~\cite{zelle-1996-geoquery,zettlemoyer-2005-ccg}.  For example, the phrase ``Who is the
president of the United States?'' might be mapped onto the query
$\lambda(x).$\formalpredicate{/government/president\_of}($x$, \formalpredicate{USA}), which, when
executed against Freebase~\cite{freebase-2008-bollacker}, returns \formalpredicate{Barack Obama}.
By mapping phrases to executable statements, semantic parsers can leverage large, curated sources
of knowledge to answer questions~\cite{berant-2013-semantic-parsing-qa}.

This benefit comes with an inherent limitation, however---semantic parsers can only produce
executable statements within their manually-produced schema.  There is no query against Freebase
that can answer questions like ``Who are the Democratic front-runners in the US election?'', as
Freebase does not encode information about front-runners.  Semantic parsers trained for Freebase
fail on these kinds of questions.

To overcome this limitation, recent work has proposed methods for \emph{open vocabulary semantic
parsing}, which replace a formal KB with a probabilistic database learned from a text corpus.  In
these methods, language is mapped onto queries with predicates derived directly from the text
itself~\cite{lewis-2013-combined-distributional-and-logical-semantics,%
krishnamurthy-2015-semparse-open-vocabulary}.  For instance, the question above might be mapped to
$\lambda(x).$\lexicalpredicate{president\_of}($x$, \formalpredicate{USA}).  This query is not
executable against any KB, however, and so open vocabulary semantic parsers must \emph{learn}
execution models for the predicates found in the text.  They do this with a distributional approach
similar to word embedding methods, giving them broad coverage, but lacking access to the large,
curated KBs available to traditional semantic parsers.

\begin{figure*}[ht]
  \small
  \centering
  \ovalbox{\begin{minipage}{.95\linewidth}
    \centering
    ~~
    \begin{minipage}{0.18\linewidth}
      \textbf{Input Text}\\Italian architect \blank{}
    \end{minipage}
    ~~ $\longrightarrow$ ~~~~~~
    \begin{minipage}{0.7\linewidth}
      \textbf{Logical Form}\\
      $\lambda x.\lexicalpredicate{architect}(x)~\land~\lexicalpredicate{architect\_N/N}(\entity{Italy}, x)$
    \end{minipage}

    \vspace{.1in}
    ~~
    \begin{minipage}{0.18\linewidth}
      \textbf{Candidate entities}\\ \entity{Palladio}\\ \entity{Obama}\\ $\cdots$
    \end{minipage}
    ~~ $\longrightarrow$ ~~~~~~
    \begin{minipage}{0.7\linewidth}
      \textbf{Probability that entity is in denotation}\\
      $\prob(\lexicalpredicate{architect}(\entity{Palladio}))
      \times
      \prob(\lexicalpredicate{architect\_N/N}(\entity{Italy}, \entity{Palladio})) = 0.79$\\
      $\prob(\lexicalpredicate{architect}\tikzmark{enda}(\entity{Obama}))
      \times
      \prob(\lexicalpredicate{architect\_N/N}\tikzmark{endb}(\entity{Italy}, \entity{Obama})) = 0.01$\\
      $\cdots$
      \vspace{-.05in}
    \end{minipage}
  \end{minipage}}

  \vspace{.05in}

  \centering
  \ovalbox{\begin{minipage}{0.4\linewidth}
    \centering
    \textbf{Category \tikzmark{starta} models:}

    {\small Predicate parameters for \lexicalpredicate{architect}:}\\
    \begin{tabular}{@{}ll}
      ~~~$\theta$: &[.2, -.6, \ldots] \\
      ~~~$\omega$: & \textsc{type:architect} $\rightarrow$ .52 \\
                & \textsc{type:designer} $\rightarrow$ .32 \\
                & \textsc{nationality:Italy} $\rightarrow$ .20 \\
                & $\cdots$
    \end{tabular}

    {\small Entity parameters for \entity{Palladio}:}\\
    \begin{tabular}{@{}ll}
      ~~~$\phi$: &[.15, -.8, \ldots] \\
      ~~~$\psi$: & \textsc{type:architect} $\rightarrow$ 1 \\
              & \textsc{nationality:Italy} $\rightarrow$ 1 \\
              & $\cdots$ \\
    \end{tabular}

    $\prob(\lexicalpredicate{architect}(\entity{Palladio})) = \sigma(\theta^T\phi + \omega^T\psi)$
  \end{minipage}}
  ~~~~
  \ovalbox{\begin{minipage}{0.5\linewidth}
    \centering
    \textbf{Relation \tikzmark{startb} models:}

    {\small Predicate parameters for \lexicalpredicate{architect\_N/N}:}\\
    \begin{tabular}{@{}ll}
      ~~~$\theta$: &[-.9, .1, \ldots] \\
      ~~~$\omega$: & \textsc{/person/nationality\textsuperscript{-1}} $\rightarrow$ .29 \\
                & \textsc{/structure/architect} $\rightarrow$ .11 \\
                & \textsc{/person/ethnicity\textsuperscript{-1}} $\rightarrow$ .05 \\
                & $\cdots$
    \end{tabular}

    {\small Entity pair parameters for (\entity{Italy}, \entity{Palladio}):}\\
    \begin{tabular}{@{}lll}
      ~~~$\phi$: &[-.8, .2, \ldots] \\
      ~~~$\psi$: & \textsc{/person/nationality\textsuperscript{-1}} $\rightarrow$ 1 \\
              & \textsc{/structure/architect} $\rightarrow$ 0 \\
              & $\cdots$ \\
    \end{tabular}

    $\prob(\lexicalpredicate{architect\_N/N}(\entity{Italy}, \entity{Palladio})) = \sigma(\theta^T\phi + \omega^T\psi)$
  \end{minipage}}

  \begin{tikzpicture}[remember picture,overlay]
    \draw[line width=1pt,>=stealth,to path={-| (\tikztotarget)}]
    ($ (starta.north east) + (-4pt,4pt) $) edge[-] ($ (starta.north east) + (-4pt, 9pt) $);%

    \draw[line width=1pt,>=stealth,to path={-| (\tikztotarget)}]
    ($ (starta.north east) + (-4pt,9pt) $) edge[->] ( $ (enda.north west) + (-4pt,-5pt) $ );%

    \draw[line width=1pt,>=stealth,to path={-| (\tikztotarget)}]
    ($ (startb.north east) + (-4pt,4pt) $) edge[-] ($ (startb.north east) + (-4pt, 9pt) $);%

    \draw[line width=1pt,>=stealth,to path={-| (\tikztotarget)}]
    ($ (startb.north east) + (-4pt,9pt) $) edge[->] ( $ (endb.north west) + (-4pt,-5pt) $ );%
  \end{tikzpicture}

  \caption{Overview of the components of our model.  Given an input text, we use a CCG parser and
  an entity linker to produce a logical form with predicates derived from the text (shown in
  italics).  For each predicate, we learn a distributional vector $\theta$, as well as weights
  $\omega$ associated with a set of selected Freebase queries.  For each entity and entity pair, we
  learn a distributional vector $\phi$, and we extract a binary feature vector $\psi$ from
  Freebase, indicating whether each entity or entity pair is in the set returned by the selected
  Freebase queries.  These models are combined to assign probabilities to candidate entities.}
  \label{fig:overview}
\end{figure*}
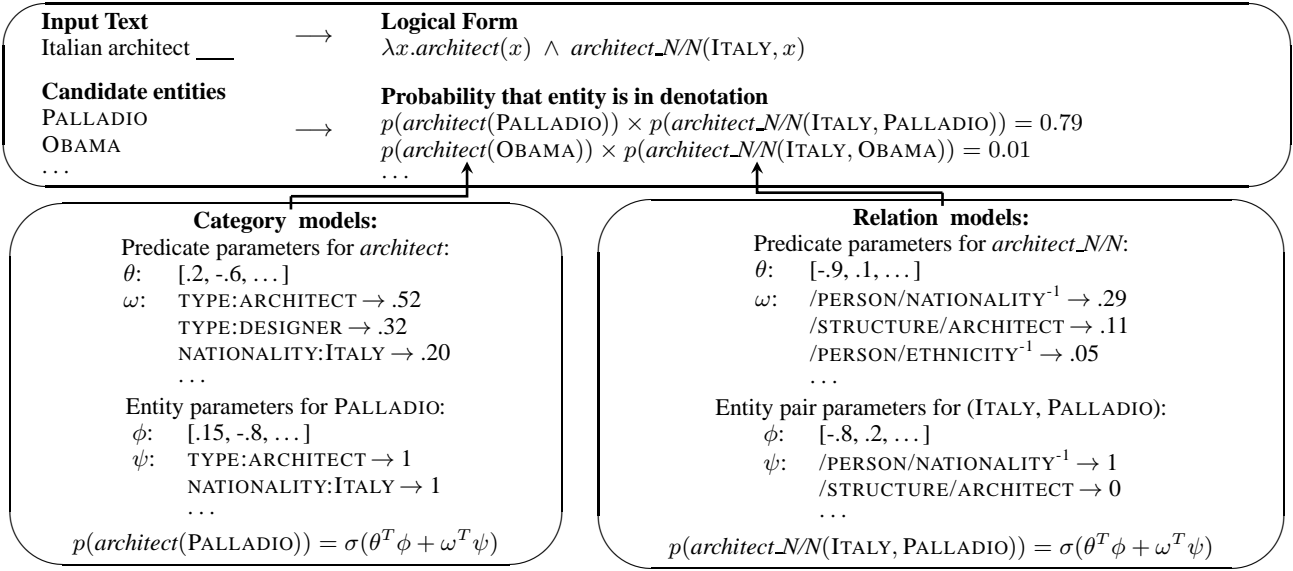

Prior work in semantic parsing, then, has either had direct access to the information in a
knowledge base, or broad coverage over all of natural language using the information in a large
corpus, but not both.

In this work, we show how to combine these two approaches by incorporating KB information into open
vocabulary semantic parsing models.  Our key insight is that formal KB queries can be converted
into \emph{features} that can be added to the learned execution models of open vocabulary semantic
parsers.  This conversion allows open vocabulary models to use the KB fact
\formalpredicate{/government/president\_of}(\formalpredicate{BarackObama}, \formalpredicate{USA})
when scoring \lexicalpredicate{president\_of}(\formalpredicate{BarackObama},
\formalpredicate{USA}), without requiring the model to map the language onto a single formal
statement.  Crucially, this featurization also allows the model to use these KB facts even when
they only provide \emph{partial} information about the language being modeled.  For example,
knowing that an entity is a \formalpredicate{politician} is very helpful information for deciding
whether that entity is a front-runner.  Our approach, outlined in \figref{overview}, effectively
learns the meaning of a word as a distributional vector plus a \emph{weighted combination of
Freebase queries}, a considerably more expressive representation than those used by prior work.

While this combination is the main contribution of our work, we also present some small
improvements that allow open vocabulary semantic parsing models to make better use of KB
information when it is available: improving the logical forms generated by the semantic parser, and
employing a simple technique from related work for generating candidate entities from the KB.

We demonstrate our approach on the task of answering open-domain fill-in-the-blank natural language
questions.  By giving open vocabulary semantic parsers direct access to KB information, we improve
mean average precision on this task by over 120\%.

\section{Open vocabulary semantic parsing}
\label{sec:jayant-semparse}

In this section, we briefly describe the current state-of-the-art model for open vocabulary
semantic parsing, introduced by Krishnamurthy and
Mitchell~\shortcite{krishnamurthy-2015-semparse-open-vocabulary}.  Instead of mapping text to
Freebase queries, as done by a traditional semantic parser, their method parses text to a
\emph{surface logical form} with predicates derived directly from the words in the text (see
\figref{overview}).  Next, a distribution over denotations for each predicate is learned using a
matrix factorization approach similar to that of Riedel et
al.~\shortcite{riedel-2013-mf-universal-schema}.  This distribution is concisely represented using
a probabilistic database, which also enables efficient probabilistic execution of logical form
queries.

The matrix factorization has two sets of parameters: each category or relation has a learned
$k$-dimensional embedding $\theta$, and each entity or entity pair has a learned $k$-dimensional
embedding $\phi$. The probability assigned to a category instance $c(e)$ or relation instance
$r(e_1, e_2)$ is given by:
\begin{align*} \prob(c(e)) &= \sigma ( \theta_c^T \phi_e ) \\ \prob(r(e_1, e_2)) &= \sigma (
  \theta_r^T \phi_{(e_1, e_2)} ) \end{align*}

The probability of a predicate instance is the sigmoided inner product of the corresponding
predicate and entity embeddings.  Predicates with nearby embeddings will have similar distributions
over the entities in their denotation. The parameters $\theta$ and $\phi$ are learned using a query
ranking objective that optimizes them to rank entities observed in the denotation of a logical form
above unobserved entities.  Given the trained predicate and entity parameters, the system is
capable of efficiently computing the marginal probability that an entity is an element of a logical
form's denotation using approximate inference algorithms for probabilistic databases.

The model presented in this section is purely \emph{distributional}, with predicate and entity
models that draw only on co-occurrence information found in a corpus.  In the following sections,
we show how to augment this model with information contained in large, curated KBs such as
Freebase.

\section{Converting Freebase queries to features}
\label{sec:queries-as-features}

Our key insight is that the executable queries used by traditional semantic parsers can be
converted into features that provide KB information to the execution models of open vocabulary
semantic parsers.  Here we show how this is done.

Traditional semantic parsers map words onto distributions over executable queries, select one to
execute, and return sets of entities or entity pairs from a KB as a result.  Instead of executing a
single query, we can simply execute \emph{all} possible queries and use an entity's (or entity
pair's) membership in each set as a feature in our predicate models.

There are two problems with this approach: (1) the set of all possible queries is intractably
large, so we need a mechanism similar to a semantic parser's lexicon to select a small set of
queries for each word; and (2) executing hundreds or thousands of queries at runtime for each
predicate and entity is not computationally tractable.  To solve these problems, we use a
graph-based technique called subgraph feature extraction (SFE)~\cite{gardner-2015-sfe}.

\subsection{Subgraph feature extraction}

SFE is a technique for generating feature matrices over node pairs in graphs with labeled edges.
When the graph corresponds to a formal KB such as Freebase, the features generated by SFE are
isomorphic to statements in the KB schema~\cite{gardner-2015-thesis}.  This means that we can use
SFE to generate a feature vector for each entity (or entity pair) which succinctly captures the set
of all statements\footnote{In a restricted class, which contains Horn clauses and a few other
things; see Gardner~\shortcite{gardner-2015-sfe} for more details.} in whose denotations the entity
(or entity pair) appears.  Using this feature vector as part of the semantic parser's entity models
solves problem (2) above, and performing feature selection for each predicate solves problem (1).

\begin{figure}
  {\center
  \begin{tikzpicture}[
    ->,
    shorten >=1pt,
    auto,
    node distance=3cm,
    thick,
    main node/.style={draw=none,fill=none}
    ]

    \node[main node] (1) {\entity{Palladio}};
    \node[main node] (2) [right=4cm of 1] {\entity{Italy}};
    \node[main node] (3) [below=1.7cm of 1] {\entity{architect}};
    \node[main node] (4) [below=1.7cm of 2] {\entity{country}};
    \node[main node] (5) [below right=.1cm and .5cm of 3] {\entity{Villa Capra}};

    \path[]
      (1) edge node [sloped, anchor=center, above] {\formalpredicate{nationality}} (2)
      (1) edge node [sloped, anchor=center, above] {\formalpredicate{type}} (3)
      (1) edge node [sloped, anchor=center, above] {\formalpredicate{designed}} (5)
      (2) edge node [sloped, anchor=center, above] {\formalpredicate{type}} (4)
      (5) edge node [sloped, anchor=center, above] {\formalpredicate{located\_in}} (2);
  \end{tikzpicture}
  }

  \textbf{Features between \entity{Palladio} and \entity{Italy}}:\\
  \pathstart\formalpredicate{nationality}\pathend\\
  \pathstart\formalpredicate{designed}$\rightarrow$\formalpredicate{located\_in}\pathend

  \textbf{Features for \entity{Palladio}}:\\
  \pathstart\formalpredicate{nationality}\pathend\\
  \pathstart\formalpredicate{nationality}\pathend:\entity{Italy}\\
  \pathstart\formalpredicate{type}\pathend:\entity{Architect}\\
  \pathstart\formalpredicate{designed}$\rightarrow$\formalpredicate{located\_in}\pathend \\
  \pathstart\formalpredicate{designed}$\rightarrow$\formalpredicate{located\_in}\pathend:\entity{Italy}

  \caption{A subset of the Freebase graph, and some example extracted features.  The actual
  Freebase relations and entity identifiers used are modified here to aid readability.}
  \label{fig:sfe}
\end{figure}

Some example features extracted by SFE are shown in \figref{sfe}.  For entity pairs, these features
include the sequence of edges (or \emph{paths}) connecting the nodes corresponding to the entity
pair.  For entities, these features include the set of paths connected to the node, optionally
including the node at the end of the path.  Note the correspondence between these features and
Freebase queries: the path
\pathstart\formalpredicate{designed}$\rightarrow$\formalpredicate{located\_in}\pathend{} can be
executed as a query against Freebase, returning a set of (architect, location) entity pairs, where
the architect designed a structure in the location. (\entity{Palladio}, \entity{Italy}) is one such
entity pair, so this pair has a feature value of 1 for this query.

\subsection{Feature selection}
\label{sec:feature-selection}

The feature vectors produced by SFE contain tens of millions of possible formal statements.  Out of
these tens of millions of formal statements, only a handful represent relevant Freebase queries for
any particular predicate.  We therefore select a small number of statements to consider for each
learned predicate in the open vocabulary semantic parser.

We select features by first summing the entity and entity pair feature vectors seen with each
predicate in the training data. For example, the phrase ``Italian architect Andrea Palladio'' is
considered a positive training example for the predicate instances
$\lexicalpredicate{architect}(\entity{Palladio})$ and
$\lexicalpredicate{architect\_N/N}(\entity{Italy}, \entity{Palladio})$.  We add the feature vectors
for \entity{Palladio} and (\entity{Italy}, \entity{Palladio}) to the feature counts for the
predicates \lexicalpredicate{architect} and \lexicalpredicate{architect\_N/N}, respectively.  This
gives a set of counts \formalpredicate{count}($\pi$), \formalpredicate{count}($f$), and
\formalpredicate{count}($\pi\land f$), for each predicate $\pi$ and feature $f$.  The features are
then ranked by PMI for each predicate by computing $\frac{\formalpredicate{count}(\pi\land
f)}{\formalpredicate{count}(\pi)\formalpredicate{count}(f)}$.  After removing low-frequency
features, we pick the $k=100$ features with the highest PMI values for each predicate to use in our
model.

\section{Combined predicate models}
\label{sec:method}

Here we present our approach to incorporating KB information into open vocabulary semantic parsers.
Having described how we use SFE to generate features corresponding to statements in a formal
schema, adding these features to the models described in \secref{jayant-semparse} is
straightforward.

We saw in \secref{jayant-semparse} that open vocabulary semantic parsers learn distributional
vectors for each category, relation, entity and entity pair.  We augment these vectors with the
feature vectors described in \secref{queries-as-features}.  Each category and relation receives a
weight $\omega$ for each selected Freebase query, and each entity and entity pair has an associated
feature vector $\psi$.  The truth probability of a category instance $c(e)$ or relation instance
$r(e_1, e_2)$ is thus given by:
\begin{align*}
  \prob(c(e)) &= \sigma ( \theta_c^T \phi_e + \omega_c^T \psi_c(e)) \\
  \prob(r(e_1, e_2)) &= \sigma ( \theta_r^T \phi_{(e_1, e_2)} + \omega_r^T \psi_r(e_1, e_2) )
\end{align*}

In these equations, $\theta$ and $\phi$ are learned predicate and entity embeddings, as described
in \secref{jayant-semparse}. The second term in the sum represents our new features and their
learned weights.  $\psi_c(e)$ and $\psi_r(e_1, e_2)$ are SFE feature vectors for each entity and
entity pair; a different set of features is chosen for each predicate $c$ and $r$, as described in
\secref{feature-selection}.  $\omega_c$ and $\omega_r$ are learned weights for these features.

In our model, there are now three sets of parameters to be learned: (1) $\theta$, low-dimensional
distributional vectors trained for each predicate; (2) $\phi$, low-dimensional distributional
vectors trained for each entity and entity pair; and (3) $\omega$, weights associated with the
selected formal SFE features for each predicate.  All of these parameters are optimized jointly,
using the same method described in \secref{jayant-semparse}.

Note here that each SFE feature corresponds to a query over the formal schema, defining a set of
entities (or entity pairs).  The associated feature weight measures the likelihood that an entity
in this set is also in the denotation of the surface predicate. Our models include \emph{many} such
features for each surface predicate, effectively mapping each surface predicate onto a
\emph{weighted combination of Freebase queries}.

\section{Making full use of KB information}

In addition to improving predicate models, as just described, adding KB information to open
vocabulary semantic parsers suggests two other simple improvements: (1) using more specific logical
forms, and (2) generating candidate entities from the KB.

\subsection{Logical form generation}
\label{sec:better-lfs}

Krishnamurthy and Mitchell~\shortcite{krishnamurthy-2015-semparse-open-vocabulary} generate logical
forms from natural language statements by computing a syntactic CCG parse, then applying a
collection of rules to produce logical forms. However, their logical form analyses do not model
noun-mediated relations well. For example, given the phrase ``Italian architect Andrea Palladio,''
their system's logical form would include the relation $\lexicalpredicate{N/N}(\entity{Italy},
\entity{Palladio})$. Here, the \lexicalpredicate{N/N} predicate represents a generic noun modifier
relation; however, this relation is too vague for the predicate model to accurately learn its
denotation. A similar problem occurs with prepositions and possessives, e.g., it is similarly hard
to learn the denotation of the predicate \lexicalpredicate{of}.

Our system improves the analysis of noun-mediated relations by simply including the noun in the
predicate name. In the architect example above, our system produces the relation
\lexicalpredicate{architect\_N/N}. It does this by concatenating all intervening noun modifiers
between two entity mentions and including them in the predicate name; for example, ``Illinois
attorney general Lisa Madigan'' produces the predicate \lexicalpredicate{attorney\_general\_N/N}.
We similarly improve the analyses of prepositions and possessives to include the head noun. For
example, ``Barack Obama, president of the U.S.'' produces the predicate instance
$\lexicalpredicate{president\_of}(\entity{Barack Obama}, \entity{U.S.})$, and ``Rome, Italy's
capital'' produces the predicate \lexicalpredicate{'s\_capital}. This process generates more
specific predicates that more closely align with the KB facts that we make available to the
predicate models.

\subsection{Candidate entity generation}
\label{sec:better-candidates}

A key benefit of our predicate models is that they are able to assign scores to entity pairs that
were never seen in the training data. Distributional models have no learned vectors for these
entity pairs and therefore assume $\prob(r(e_1,e_2)) = 0$ for unseen entity pairs $(e_1,e_2)$. This
limits the recall of these models when applied to question answering, as entity pairs will not have
been observed for many correct, but rare entity answers. In contrast, because our models have
access to a large KB, the formal component of the model can always give a score to any entity pair
in the KB.  This allows our model to considerably improve question answering performance on rare
entities.

It would be computationally intractable to consider \emph{all} Freebase entities as answers to
queries, and so we use a simple candidate entity generation technique to consider only a small set
of likely entities for a given query.  We first find all entities in the query, and consider as
candidates any entity that has either been seen at training time with a query entity or is directly
connected to a query entity in Freebase.\footnote{Or connected by a mediator node, which is how
Freebase represents relations with more than two arguments.} This candidate entity generation is
common practice for recent question answering models over
Freebase~\cite{yih-2015-semparse-query-graph}, though, for the reasons stated above, it has not
been used previously in open vocabulary semantic parsing models.

\section{Evaluation}
\label{sec:evaluation}

We evaluate our open-vocabulary semantic parser on a fill-in-the-blank natural language query task.
Each test example is a natural language phrase containing at least two Freebase entities, one of
which is held out.  The system must propose a ranked list of Freebase entities to fill in the blank
left by the held out entity, and the predicted entities are then judged manually for correctness.
We compare our proposed models, which combine distributional and formal elements, with a purely
distributional baseline from prior work.  All of the data and code used in these experiments is
available at http://github.com/allenai/open\_vocab\_semparse.

\subsection{Data}

Much recent work on semantic parsing has been evaluated using the WebQuestions
dataset~\cite{berant-2013-semantic-parsing-qa}.  This dataset is not suitable for evaluating our
model because it was filtered to only questions that are mappable to Freebase queries.  In
contrast, our focus is on language that is \emph{not} directly mappable to Freebase.  We thus use
the dataset introduced by Krishnamurthy and
Mitchell~\shortcite{krishnamurthy-2015-semparse-open-vocabulary}, which consists of the ClueWeb09
web corpus\footnote{http://www.lemuproject.org/clueweb09.php} along with Google's FACC entity
linking of that corpus to Freebase~\cite{gabrilovich-2013-clueweb-entity-linking}.  For training
data, 3 million webpages from this corpus were processed with a CCG parser to produce logical
forms~\cite{krishnamurthy-2014-joint-ccg}.  This produced 2.1m predicate instances involving 142k
entity pairs and 184k entities.  After removing infrequently-seen predicates (seen fewer than 6
times), there were 25k categories and 4.2k relations.

We also used the test set created by Krishnamurthy and Mitchell, which contains 220 queries
generated in the same fashion as the training data from a separate section of ClueWeb.  However, as
they did not release a development set with their data, we used this set as a development set.  For
a final evaluation, we generated another, similar test set from a different held out section of
ClueWeb, in the same fashion as done by Krishnamurthy and Mitchell.  This final test set contains
307 queries.

\subsection{Models}

We compare three models in our experiments: (1) the distributional model of Krishnamurthy and
Mitchell, described in \secref{jayant-semparse}, which is the current state-of-the-art method for
open vocabulary semantic parsing; (2) a formal model (new to this work), where the distributional
parameters $\theta$ and $\phi$ in \secref{method} are fixed at zero; and (3) the combined model
described in \secref{method} (also new to this work).  In each of these models, we used vectors of
size 300 for all embeddings.  Except where noted, all experiments use our modified logical forms
(\secref{better-lfs}) and our entity proposal mechanism (\secref{better-candidates}).  We do not
compare against any traditional semantic parsers, as more than half of the questions in our dataset
are not answerable by Freebase queries, and so are out of scope for those
parsers~\cite{krishnamurthy-2015-semparse-open-vocabulary}.

\subsection{Methodology}

Given a fill-in-the-blank query such as ``Italian architect \blank{}'', each system produces a
ranked list of 100 candidate entities.  To compare the output of the systems, we follow a pooled
evaluation protocol commonly used in relation extraction and information
retrieval~\cite{west-2014-kbc-via-qa,riedel-2013-mf-universal-schema}.  We take the top 30
predictions from each system and manually annotate whether they are correct, and use those
annotations to compute the average precision (AP) and reciprocal rank (RR) of each system on the
query.  Average precision is defined as $\frac{1}{m}\sum^m_{k=1} \mathrm{Prec}(k) \times
\mathrm{Correct}(k)$, where $\mathrm{Prec}(k)$ is the precision at rank $k$, $\mathrm{Correct}(k)$
is an indicator function for whether the $k$th answer is correct, and $m$ is number of returned
answers (up to 100 in this evaluation).  AP is equivalent to calculating the area under a
precision-recall curve.  Reciprocal rank is computed by first finding the rank $r$ of the first
correct prediction made by a system.  Reciprocal rank is then $\frac{1}{r}$, ranging from 1 (if the
first prediction is correct) to 0 (if there is no correct answer returned).  In the tables below we
report \emph{mean} average precision (MAP) and \emph{mean} reciprocal rank (MRR), averaged over all
of the queries in the test set.  We also report a weighted version of MAP, where the AP of each
query is scaled by the number of annotated correct answers to the query (shown as W-MAP in the
tables for space considerations).

\subsection{Results}

\begin{table}
  \centering
  {\small
    \begin{tabular}{l@{\hskip 0pt}ccc}
      \toprule
      Model & K\&M's LFs & Our LFs & Delta \\
      \midrule
      Distributional (K\&M 2015) & .269 & \textbf{.284} & +.015 \\
      Formal & .231 & \textbf{.276} & +.045 \\
      Combined & .313 & \textbf{.335} & +.022 \\
      \bottomrule
    \end{tabular}
  }
  \caption{Improvement in mean average precision when using our logical forms on the development
  set.}
  \label{tab:better-lfs}
\end{table}

We first show the effect of the new logical forms introduced in \secref{better-lfs}.  As can be
seen in \tabref{better-lfs}, with our improved logical forms, all models are better able to capture
the semantics of language.  This improvement is most pronounced in the formal models, which have
more capacity to get specific features from Freebase with the new logical forms.  As our logical
forms give all models better performance, the remaining experiments we present all use these
logical forms.

\begin{table}
  \centering
  {\small
    \begin{tabular}{lccc}
      \toprule
      Model & MAP & W-MAP & MRR \\
      \midrule
      Distributional (K\&M 2015) & .163 & .163 & .288 \\
      With freebase candidates & .\textbf{229} & \textbf{.275} & \textbf{.312} \\
      \midrule
      Relative improvement & 40\% & 69\% & 8\% \\
      \bottomrule
    \end{tabular}
  }
  \caption{Improvement to the distributional model when using our candidate entity generation.}
  \label{tab:better-candidates}
\end{table}

We next show the improvement gained by using the simple candidate entity generation outlined in
\secref{better-candidates}.  By simply appending the list of connected entities in Freebase to the
end of the rankings returned by the distributional model, MAP improves by 40\% (see
\tabref{better-candidates}).  The connectedness of an entity pair in Freebase is very informative,
especially for rare entities that are not seen together during training.

\begin{table}
  \centering
  {\small
    \begin{tabular}{lccc}
      \toprule
      Model & MAP & W-MAP & MRR \\
      \midrule
      Distributional (K\&M 2015) & .284 & .371 & .379 \\
      Formal & .276 & .469 & .334 \\
      Combined & \textbf{.335} & \textbf{.477} & \textbf{.429} \\
      \midrule
      Relative improvement & 18\% & 29\% & 13\% \\
      \bottomrule
    \end{tabular}
  }
  \caption{Development set results for our fill-in-the-blank task.  The combined model
  significantly improves MAP over prior work.}
  \label{tab:dev-results}
\end{table}

\tabref{dev-results} shows a comparison between the semantic parsing models on the development set.
As can be seen, the combined model significantly improves performance over prior work, giving a
relative gain in weighted MAP of 29\%.

\begin{table}
  \centering
  {\small
    \begin{tabular}{lccc}
      \toprule
      Model & MAP & W-MAP & MRR \\
      \midrule
      Distributional (K\&M 2015) & .229 & .275 & .312 \\
      Formal & .355 & .495 & .419 \\
      Combined & \textbf{.370} & \textbf{.513} & \textbf{.469} \\
      \midrule
      Relative improvement & 62\% & 87\% & 50\% \\
      \bottomrule
    \end{tabular}
  }
  \caption{Final test results set for our fill-in-the-blank task.  The combined model improves over
  prior work by 50--87\% on our metrics.  These improvements over the baseline are \emph{after} the
  baseline has been improved by the methods developed in this paper, shown in \tabref{better-lfs}
  and \tabref{better-candidates}.  The cumulative effect of the methods presented in this work is
  an improvement of over 120\% in MAP.}
  \label{tab:final-results}
\end{table}

\tabref{final-results} shows that these improvements are consistent on the final test set, as well.
The performance improvement seen by the combined model is actually larger on this set, with gains
on our metrics ranging from 50\% to 87\%.

On both of these datasets, the difference in MAP between the combined model and the distributional
model is statistically significant (by a paired permutation test, $p < 0.05$).  The differences
between the combined model and the formal model, and between the formal model and the
distributional model, are not statistically significant, as each method has certain kinds of
queries that it performs well on.  Only the combined model is able to consistently outperform the
distributional model on all kinds of queries.

\subsection{Discussion}

Our model tends to outperform the distributional model on queries containing predicates with exact
or partial correlates in Freebase. For example, our model obtains nearly perfect average precision
on the queries ``French newspaper \blank{}'' and ``Israeli prime minister \blank{},'' both of which
can be exactly expressed in Freebase.  The top features for \lexicalpredicate{newspaper}($x$) all
indicate that $x$ has type \formalpredicate{newspaper} in Freebase, and the top features for
\lexicalpredicate{newspaper\_N/N}($x$, $y$) indicate that $y$ is a newspaper, and that $x$ is
either the circulation area of $y$ or the language of $y$.

The model also performs well on queries with partial Freebase correlates, such as ``Microsoft head
honcho \blank{}'', ``The United States, \blank{}'s closest ally'', and ``Patriots linebacker
\blank{},'' although with somewhat lower average precision. The high weight features in these cases
tend to provide useful hints, even though there is no direct correlate; for example, the model
learns that ``honchos'' are people, and that they tend to be CEOs and film producers.

There are also some areas where our model can be improved. First, in some cases, the edge sequence
features used by the model are not expressive enough to identify the correct relation in Freebase.
An example of this problem is the ``linebacker'' example above, where the features for
\lexicalpredicate{linebacker\_N/N} can capture which athletes play for which teams, but not the
\emph{positions} of those athletes. Second, our model can under-perform on predicates with no close
mapping to Freebase. An example where this problem occurs is the query ``\blank{} is a NASA
mission.'' Third, there remains room to further improve the logical forms produced by the semantic
parser, specifically for multi-word expressions. One problem occurs with multi-word noun modifiers,
e.g., ``Vice president Al Gore'' is mapped to $\lexicalpredicate{vice}(\entity{Al Gore}) \land
\lexicalpredicate{president}(\entity{Al Gore})$. Another problem is that there is no back-off with
multi-word relations. For example, the predicate \lexicalpredicate{head\_honcho\_N/N} was never
seen in the training data, so it is replaced with \lexicalpredicate{unknown}; however, it would be
better to replace it with \lexicalpredicate{honcho\_N/N}, which \emph{was} seen in the training
data. Finally, although using connected entities in Freebase as additional candidates during
inference is helpful, it often over- or under-generates candidates. A more tailored, per-query
search process could improve performance.

\section{Related work}

There is an extensive literature on building semantic parsers to answer questions against a
KB~\cite{zettlemoyer-2005-ccg,berant-2013-semantic-parsing-qa,%
krishnamurthy-2012-semantic-parsing,li-2015-semantic-parsing-scfg}.  Some of this work has used
surface (or \emph{ungrounded}) logical forms as an intermediate representation, similar to our
work~\cite{kwiatkowski-2013-ontology-matching,reddy-2014-graph-matching,%
yih-2015-semparse-query-graph,reddy-2016-dep-lambda}.  The main difference between our work and
these techniques is that they map surface logical forms to a single executable Freebase query,
while we learn execution models for the surface logical forms directly, using a weighted
combination of Freebase queries as part of the model.  None of these prior works can assign meaning
to language that is not directly representable in the KB schema.

Choi, Kwiatkowski and Zettlemoyer~\shortcite{choi-2015-semantic-parsing-partial-ontologies}
presented an information extraction system that performs a semantic parse of open-domain text,
recognizing when a predicate cannot be mapped to Freebase.  However, while they recognize when a
predicate is not mappable to Freebase, they do not attempt to learn execution models for those
predicates, nor can they answer questions using those predicates.

Yao and Van Durme~\shortcite{yao-2014-info-extraction-freebase-qa} and Dong et
al.~\shortcite{dong-2015-freebase-qa-mccnn} proposed question answering models that use similar
features to those used in this work.  However, they did not produce semantic parses of language,
instead using methods that are non-compositional and do not permit complex queries.

Finally, learning probabilistic databases in an open vocabulary semantic parser has a strong
connection with KB completion.  In addition to SFE~\cite{gardner-2015-sfe}, our work draws on work
on embedding the entities and relations in a
KB~\cite{riedel-2013-mf-universal-schema,nickel-2011-rescal,%
bordes-2013-transe,nickel-2014-are,toutanova-2015-joint-text-kb-embedding}, as well as work on
graph-based methods for reasoning with
KBs~\cite{lao-2010-original-pra,gardner-2014-vector-space-pra,neelakantan-2015-rnn-kbc,%
bornea-2015-relational-path-mining}.  Our combination of embedding methods with graph-based methods
in this paper is suggestive of how one could combine the two in methods for KB completion.  Initial
work exploring this direction has already been done by Toutanova and
Chen~\shortcite{toutanova-2015-observed-vs-latent-kbc}.


\section{Conclusion}
\label{sec:conclusion}

Prior work in semantic parsing has either leveraged large knowledge bases to answer questions, or
used distributional techniques to gain broad coverage over all of natural language.  In this paper,
we have shown how to gain both of these benefits by converting the queries generated by traditional
semantic parsers into features which are then used in open vocabulary semantic parsing models.  We
presented a technique to do this conversion in a way that is scalable using graph-based feature
extraction methods.  Our combined model achieved relative gains of over 50\% in mean average
precision and mean reciprocal rank versus a purely distributional approach.  We also introduced a
better mapping from surface text to logical forms, and a simple method for using a KB to find
candidate entities during inference.  Taken together, the methods introduced in this paper improved
mean average precision on our task from .163 to .370, a 127\% relative improvement over prior work.

This work suggests a new direction for semantic parsing research.  Existing semantic parsers map
language to a single KB query, an approach that successfully leverages a KB's predicate instances,
but is fundamentally limited by its schema. In contrast, our approach maps language to a
\emph{weighted combination of queries} plus a distributional component; this approach is capable of
representing a much broader class of concepts while still using the KB when it is helpful.
Furthermore, it is capable of using the KB even when the meaning of the language cannot be exactly
represented by a KB predicate, which is a common occurrence. We believe that this kind of approach
could significantly expand the applicability of semantic parsing techniques to more complex domains
where the assumptions of traditional techniques are too limiting.  We are actively exploring
applying these techniques to science question answering~\cite{clark-2013-kb-requirements-for-qa},
for example, where existing KBs provide only partial coverage of the questions.


\small
\bibliography{bib}
\bibliographystyle{aaai}

\end{document}